\documentclass[10pt,twocolumn,letterpaper]{article}

\usepackage[pagenumbers]{cvpr}   

\usepackage{cuted} 

\usepackage{currfile} 

\usepackage{caption} 
\usepackage{multirow}
\usepackage{colortbl}
\usepackage{makecell}
\usepackage{tabularx}
\newcolumntype{Y}{>{\centering\arraybackslash}X}
\usepackage{nicefrac}
\usepackage{tikz}
\usetikzlibrary{arrows.meta,positioning,shapes.geometric,fit,backgrounds}

\newcommand{\method}{PhysMoE\xspace}

\definecolor{bestblue}{RGB}{210, 222, 239}
\definecolor{secondblue}{RGB}{234, 239, 247}
\newcommand{\first}[1]{\cellcolor{bestblue}#1}
\newcommand{\second}[1]{\cellcolor{secondblue}#1}
\newcommand{\firstbox}{\colorbox{bestblue}{\rule{0pt}{1.1ex}\hspace{1.1ex}}}
\newcommand{\secondbox}{\colorbox{secondblue}{\rule{0pt}{1.1ex}\hspace{1.1ex}}}

\definecolor{cvprblue}{rgb}{0.21,0.49,0.74}
\usepackage[pagebackref,breaklinks,colorlinks,allcolors=cvprblue]{hyperref}

\def\confName{arXiv}
\def\confYear{2026}

\providecommand{\method}{PhysMoE}

\title{\method: Fast Spatial Grounding with Single-Token Coordinate Representation in Modality-Decoupled VLMs}

\author{
  Ruimeng Liu$^{1}$, \quad
  Denan Liang$^{1}$, \quad
  Kuan Xu$^{1}$, \quad
  Ruofei Bai$^{1}$, \quad
  Fuming Zhang$^{1}$, \\
  Shenghai Yuan$^{1}$, \quad
  Lihua Xie$^{1}$, \textit{Fellow, IEEE}
}

\makeatletter
\patchcmd{\@maketitle}{\vskip .375in}{\vskip -.22in}{}{}       
\patchcmd{\@maketitle}{\vspace*{24pt}}{\vspace*{16pt}}{}{}    
\patchcmd{\@maketitle}{\vspace*{12pt}}{\vspace*{-14pt}}{}{}    
\makeatother

\begin{document}
\maketitle

\footnotetext[1]{Center for Advanced Robotics Technology Innovation (CARTIN), School
of Electrical and Electronic Engineering, Nanyang Technological University,
Singapore 639798.}

\begin{strip}
\vspace{-16pt}
\centering
\includegraphics[width=\textwidth]{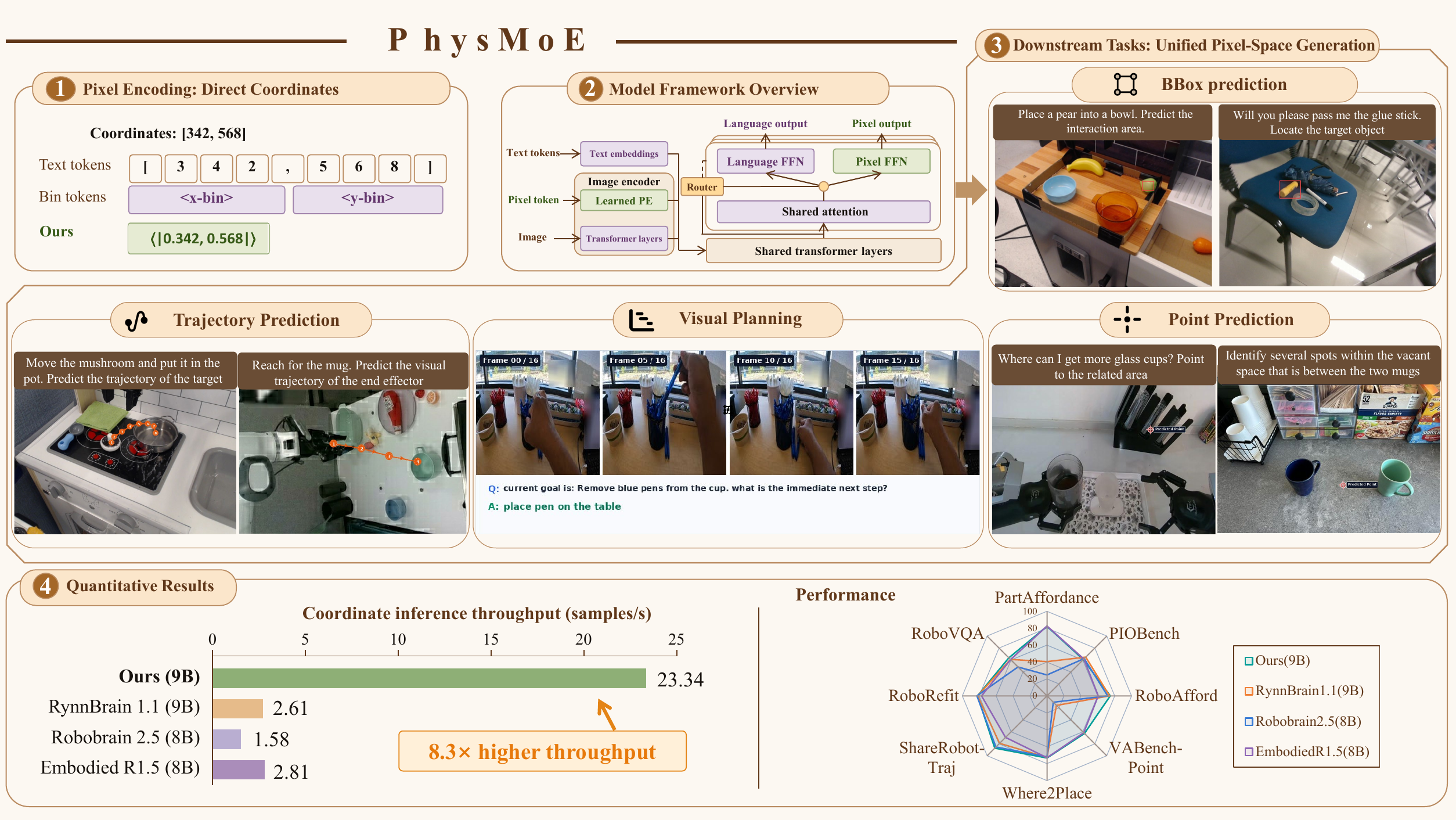}
\captionsetup{hypcap=false}
\captionof{figure}{\textbf{Overview of the \method framework}. The figure illustrates (1) direct continuous coordinate encoding, (2) the model overview with routed language and pixel FFNs, (3) unified pixel-space generation across downstream tasks (bounding box, trajectory, visual planning, and point prediction), and (4) quantitative results on coordinate inference throughput and benchmark performance.}
\captionsetup{hypcap=true}
\label{fig:teaser}
\end{strip}

\begin{abstract}
Embodied Vision-Language Models (VLMs) are backbones for robotic planning and manipulation. 
However, for coordinate prediction in grounding tasks, existing architectures suffer from two distinct bottlenecks: discrete tokenization disrupts spatial continuity with high decoding latency, while standard point regression fails on multi-contact affordances and irregularly shaped targets. 
To resolve these challenges, we propose \textbf{\method}, introducing three core contributions: 
(1) sharing lower layers for multimodal context while routing upper layers to specialized spatial (pixel) and linguistic (text) experts by physical modality, eliminating gradient interference; 
(2) encoding each coordinate as a single token by reusing the visual encoder's learned positional embeddings, injecting 2D spatial priors to natively align coordinates with visual features and ease training; and 
(3) differentiable Signed Distance Field (SDF) supervision on pointing, accommodating affordances and irregular geometries without costly mask decoders. 
Compared to base models, \method achieves a 15.7\% average gain and an $8.31\times$ throughput speedup across 9 coordinate-based grounding benchmarks (trajectory, affordance, and localization), while preserving multimodal understanding and reasoning across 11 text-based embodied benchmarks. Our work is open-sourced at \url{https://github.com/lrm20011/ps_moe_open}.
\end{abstract}
    
\section{Introduction}
\label{sec:intro}

Embodied Vision-Language Models (VLMs) increasingly connect visual
understanding and language-based reasoning with robotic
tasks~\cite{zitkovich2023rt2,geminirobotics2025,robobrain2024,rynnbrain2024}.
Beyond interpreting instructions and scenes, they must determine
\emph{where} to act: grasping contact points, object placement locations,
or image-space end-effector
waypoints~\cite{yuan2024robopoint,yuan2026fromseeing,lee2025molmoact,rynnbrain2024}.
These spatial outputs link task-level reasoning to downstream planning
and control, requiring geometrically valid predictions with low generation latency. 
The central problem is therefore \emph{how to integrate accurate
and efficient spatial grounding into an embodied VLM while preserving
its language understanding and reasoning capabilities}.

\textbf{Existing} approaches provide different mechanisms for connecting VLMs with
spatial outputs. A common strategy represents coordinates as numerical
strings, such as \texttt{"[342, 518]"},
or quantized position
tokens~\cite{bai2025qwen3vl,chen2023shikra,deitke2024molmo,peng2023kosmos2,chen2021pix2seq},
allowing spatial predictions to reuse the standard autoregressive language
generation interface. Mask-based approaches instead represent target
regions explicitly through segmentation
decoders~\cite{lai2024lisa,yao2026qwen3vlseg,kirillov2023segment}.
Meanwhile, action experts in vision-language-action
models~\cite{black2024pi0} and modality-specific experts in multimodal
models~\cite{wang2024cogvlm,liang2024mot} introduce specialized computation
for action generation or multimodal inputs. These approaches offer useful
mechanisms for spatial representation and modality specialization.
Our focus is on a complementary problem: supporting continuous image-space
coordinates alongside language within a single VLM. This requires considering
not only how coordinates are represented, but also how spatial prediction
is learned jointly with language and how geometrically valid outputs are
supervised.

Three \textbf{challenges} arise in meeting these requirements.
\textbf{First, efficient and spatially grounded representation.}
Serializing a coordinate into numerical and delimiter tokens requires
multiple decoding steps, while neither numerical strings nor discrete
position identifiers inherently establish correspondence with the visual
encoder's spatial representation. A suitable representation should support
compact prediction and encode coordinates in relation to image locations.
\textbf{Second, compatible spatial--language learning.}
Introducing continuous coordinate regression alongside language generation
creates distinct prediction objectives. The model must learn precise spatial
outputs without allowing spatial specialization to compromise its pretrained
language capabilities.
\textbf{Third, supervision of geometrically valid alternatives.}
For affordance and placement tasks, the desired output may be any suitable
point within a valid region rather than a unique annotated coordinate.
Single-point supervision can penalize alternative valid predictions, and
the centroid of a non-convex region may even lie outside that region
(Fig.~\ref{fig:semimoe_arch}e). Spatial supervision should therefore distinguish
between tasks with fixed coordinate targets and those admitting a set of
valid locations.

To \textbf{address} these challenges, we propose \textbf{\method}, which treats
image-space coordinates as a \emph{continuous output modality} alongside
language (Fig.~\ref{fig:teaser}). Our guiding principle is to \emph{share
contextual understanding while specializing spatial representation,
computation, and supervision}. Each normalized 2D coordinate is represented
as a single token, embedded through bilinear interpolation of the visual
encoder's learned positional embeddings, and predicted by a regression
head in one decoding step. Lower decoder layers and self-attention remain
shared, while a lightweight router selects linguistic or spatial FFNs
in the top $K$ layers and the corresponding output head.
For fixed coordinate targets, we use Smooth $\ell_1$ regression.
For region-defined targets, we convert target masks into Euclidean Signed
Distance Fields (ESDFs) and apply a differentiable asymmetric loss that
penalizes exterior predictions and encourages an interior margin.
This supervision accommodates multiple valid locations without forcing
predictions toward a single reference point or requiring a mask decoder
at inference.

Our main contributions are summarized as follows:
\begin{enumerate}[leftmargin=*,noitemsep,topsep=2pt]
    \item We introduce a single-token continuous coordinate representation
    that reuses the visual encoder's learned positional embeddings and
    supports one-step 2D coordinate prediction, avoiding multi-token
    coordinate serialization.

    \item We propose \method, which combines shared contextual processing
    with modality-specific linguistic and spatial FFNs in the top $K$
    decoder layers, enabling spatial specialization while reducing direct
    parameter sharing between language generation and coordinate regression.

    \item We introduce differentiable ESDF supervision for region-defined
    pointing tasks, accommodating multiple valid locations and irregular
    target geometries without requiring a mask decoder at inference.

    \item We evaluate \method on 9 coordinate-based grounding benchmarks
    and 11 text-based embodied benchmarks. Our 9B model achieves a
    15.7\% relative improvement in average grounding score over its
    initialization backbone and $8.31\times$ the coordinate inference
    throughput of the fastest evaluated 8B--9B baseline, while maintaining
    comparable average text-based reasoning performance.
\end{enumerate}
\section{Related Work}
\label{sec:related}

\noindent\textbf{Embodied Vision-Language Models.}
Multimodal foundation models~\cite{radford2021learning,touvron2023llama,bai2025qwen3vl,achiam2023gpt4} have inspired a surge of embodied VLMs for robotic planning and control~\cite{driess2023palm,zitkovich2023rt2}. Dense models such as RoboBrain~\cite{robobrain2024}, RynnBrain~\cite{rynnbrain2024}, and Embodied-R1.5~\cite{embodiedr15_2024} advance physical reasoning, while recent architectures adopt Mixture-of-Experts (MoE)~\cite{shazeer2017outrageously,dai2024deepseekmoe,lin2024moellava} or Mixture-of-Transformers (MoT)~\cite{liang2024mot} structures to scale capacity or decouple modalities, including HY-Embodied-VLM~\cite{hyembodied2024} and Cosmos~3~\cite{agarwal2026cosmos}. However, these models either route across general data-driven clusters or still emit pixel coordinates as discrete text. In contrast, \method enforces a structural division of labor inside a single autoregressive VLM: self-attention stays shared for unified context, while top-layer FFNs isolate linguistic generation from continuous spatial regression.

\noindent\textbf{Coordinate Encodings in VLMs.}
Spatial coordinates are conventionally tokenized as character-level strings (\emph{e.g.}, \texttt{[}, \texttt{3}, \texttt{4}, \dots)~\cite{chen2023shikra,bai2025qwen3vl,xiao2024florence2} or discretized spatial bins~\cite{chen2021pix2seq,peng2023kosmos2}. Though vocabulary-compatible, such discrete encodings incur quantization error, sequence bloat, and language degradation during fine-tuning. Recently, LocateAnything~\cite{wang2026locateanything} accelerates grounding via parallel box decoding, but its box-specific design does not extend to other spatial outputs such as trajectories, and it sacrifices text-based performance. In contrast, \method regresses each coordinate as a single continuous token and encodes them by reusing the visual encoder's positional embeddings, supporting points, boxes, and trajectories alike.

\noindent\textbf{Multi-Peak Affordance Supervision.}
Physical affordance learning~\cite{gibson1979ecological,do2018affordancenet,nagarajan2019grounded} is inherently a \emph{multi-peak problem}, where an actionable surface contains multiple valid contact points. Standard single-point regression penalizes valid alternate modes or averages them into invalid regions. Inspired by implicit signed distance fields in 3D geometry~\cite{park2019deepsdf} and robotics~\cite{ratliff2009chomp,oleynikova2017voxblox}, we introduce 2D ESDF as a continuous, differentiable loss, resolving multi-peak ambiguity without costly mask decoders~\cite{kirillov2023segment}.

\section{Methodology}
\label{sec:method}

In this section, we present the architecture and training methodology of \method. We first detail the modality decoupled architecture with dynamic routing (\cref{sec:method_semimoe}) and describe the continuous pixel position representation with bilinear positional encoding (\cref{sec:method_pos_embed}). We then introduce the differentiable ESDF supervision for multi-peak target pointing (\cref{sec:method_masksdf}) and outline the two-stage training curriculum (\cref{sec:method_stages}).

\subsection{PhysMoE Decoupled Architecture}
\label{sec:method_semimoe}

\begin{figure*}[t]
    \centering
    \includegraphics[width=\textwidth]{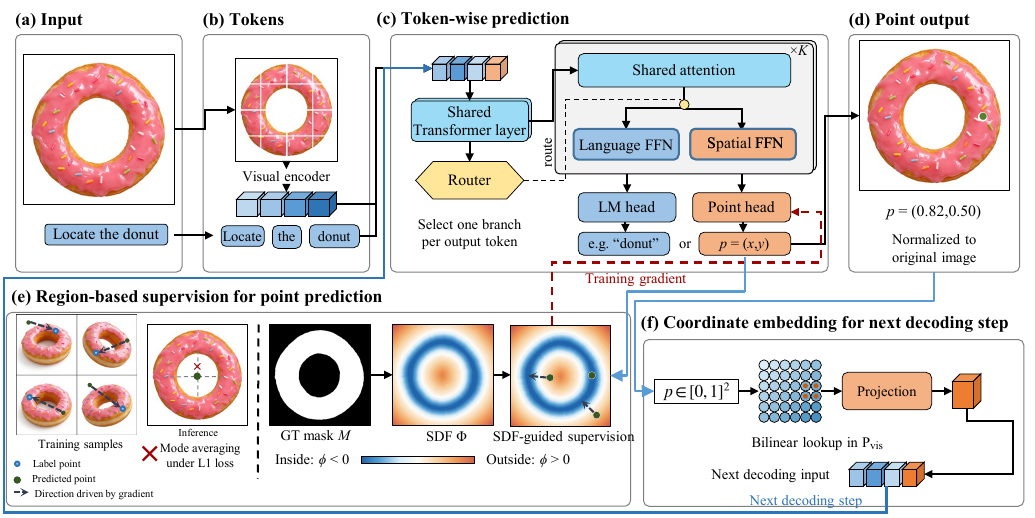}
    \caption{\textbf{\method Architecture, Region-Based Supervision, and Continuous Coordinate Embedding}. (a) Input RGB image and prompt. (b) Sequence tokenization. (c) PhysMoE layers with shared self-attention and dynamically routed language/spatial experts ($\text{FFN}_{\text{lang}}$, $\text{FFN}_{\text{pix}}$). (d) Decoupled prediction heads for language tokens and 2D coordinates. (e) Region-based Mask-SDF supervision for continuous point prediction, penalizing exterior predictions while accommodating multi-peak affordance regions without mode averaging. (f) Continuous coordinate embedding via bilinear interpolation over visual positional embeddings $\mathcal{P}_{\text{vis}}$ and linear adapter projection for subsequent decoding steps.}
    \label{fig:semimoe_arch}
\end{figure*}

Standard dense transformers process all tokens through identical Feed-Forward Networks (FFNs), causing acute gradient interference between discrete categorical text classification and continuous coordinate regression. To resolve this conflict while retaining global visual-linguistic context, we introduce the \method architecture (\cref{fig:semimoe_arch}).

\noindent\textbf{Layer Partitioning.} 
Let the Transformer backbone comprise $L$ sequential layers ($l = 0, \dots, L - 1$). We partition the decoder into two functional stages:
\begin{itemize}[leftmargin=*,noitemsep,topsep=1pt]
    \item \emph{Shared Context Layers ($l = 0, \dots, L - K - 1$)}: The initial $L - K$ layers operate as standard dense Transformer blocks where visual, textual, and spatial tokens attend to one another freely, building rich multimodal contextual representations.
    \item \emph{PhysMoE Routed Layers ($l = L - K, \dots, L - 1$)}: Across the top $K$ layers (e.g., $K = 8$), we maintain the shared self-attention mixer across all tokens to preserve uninterrupted global receptive fields. However, the subsequent feed-forward computation is decoupled into two modality-specialized experts: linguistic tokens pass through $\mathrm{FFN}_{\text{lang}}$, whereas spatial tokens pass through $\mathrm{FFN}_{\text{pix}}$, with each token's expert assigned by the dynamic token router described below. This decoupling isolates continuous spatial regression gradients from textual semantic representations.
\end{itemize}

\noindent\textbf{Dynamic Token Router.} 

At layer $L-K$, a lightweight two-layer MLP router $\mathrm{MLP}_r$ makes a single routing
decision at each position $t$ to predict the next token, determining the entire downstream path for that prediction.
We model this decision as a binary random variable $c_t \in \{0,1\}$:
linguistic tokens ($c_t = 0$) are handled by $\mathrm{FFN}_{\text{lang}}$ and the LM head,
and spatial tokens ($c_t = 1$) by $\mathrm{FFN}_{\text{pix}}$ and the pixel head.
Given the normalized hidden state $\mathbf{h}_r \in \mathbb{R}^{d_{\text{text}}}$ in the layer $L-K$, the router
outputs a vector of routing probabilities
\begin{equation}
    \mathbf{r}_t = \mathrm{softmax}\big(\mathrm{MLP}_r(\mathbf{h}_r)\big) \in [0,1]^2,
    \label{eq:router}
\end{equation}
which defines the router's conditional distribution over experts:
$P(c_t = k \mid \mathbf{h}_r) = r_{t,k}$, i.e., the $k$-th entry of $\mathbf{r}_t$ is the
predicted probability of routing token $t$ to expert $k$, with $r_{t,0} + r_{t,1} = 1$.
At inference, we apply hard gating and send each token to its most probable expert,
$\hat{c}_t = \arg\max_{k \in \{0,1\}} P(c_t = k \mid \mathbf{h}_r)$.
During training, we instead teacher-force the routing with the ground-truth modality label
$c_t^* \in \{0,1\}$, and supervise the router with a weighted cross-entropy loss:
\begin{equation}
    \mathcal{L}_{\text{router}} = - w_t \log P(c_t^* \mid \mathbf{h}_r),
    \label{eq:router_loss}
\end{equation}
where $w_t = w_{\text{eos}}$ for the end-of-sequence (EOS) token and $w_t = 1$ otherwise.

\noindent\textbf{Decoupled Dual Prediction Heads.} 
At the final layer $L$, hidden states $h_t$ are mapped to predictions via two task-specific heads; each token is decoded only by the head selected by its routing decision $c_t$:
\begin{itemize}[leftmargin=*,noitemsep,topsep=1pt]
    \item \emph{LM Head}: A single linear projection $\mathbf{W}_{\text{lm}} \in \mathbb{R}^{V \times d_{\text{text}}}$ that maps hidden states to vocabulary logits, supervised by weighted cross-entropy:
    \begin{equation}
        \mathcal{L}_{\text{LM}} = - w_t \log P(y_{t+1}^* \mid \mathbf{h}_t),
    \end{equation}
    where $P(y_{t+1} \mid \mathbf{h}_t) = \text{Softmax}(\mathbf{W}_{\text{lm}} \mathbf{h}_t)$.
    \item \emph{Pixel Head}: A two-layer MLP $\mathrm{MLP}_p$ that directly regresses continuous normalized 2D coordinates:
    \begin{equation}
        \hat{\mathbf{p}}_t = \sigma\big(\mathrm{MLP}_p(\mathbf{h}_t)\big) \in [0, 1]^2,
        \label{eq:pix_head}
    \end{equation}
    where $\sigma(\cdot)$ is the Sigmoid activation.
    For fixed coordinate targets $\mathbf{p}_t^* = (x_t^*, y_t^*)$, such as trajectory waypoints and bounding-box corners, the pixel head is supervised with the Smooth L1 loss:
    \begin{equation}
        \mathcal{L}_{\text{coord}} = \frac{1}{2} \sum_{j \in \{x, y\}} \ell_\beta\big(\hat{p}_{t,j} - p_{t,j}^*\big),
        \label{eq:coord_loss}
    \end{equation}
    \begin{equation}
        \ell_\beta(u) = \begin{cases}
            \frac{u^2}{2\beta}, & \text{if } |u| < \beta , \\
            |u| - \frac{\beta}{2}, & \text{otherwise} ,
        \end{cases}
        \label{eq:smooth_l1}
    \end{equation}
    where $\beta = 0.035$. Region targets (\emph{e.g.}, affordance masks) are instead supervised by the ESDF loss (\cref{sec:method_masksdf}).
\end{itemize}

\subsection{Continuous Coordinate Representation and Encoding}
\label{sec:method_pos_embed}

\noindent\textbf{Representation Syntax.} Within conversational dialogues and instructions, 2D coordinates are formatted as $\langle|x, y|\rangle$, where $x, y \in [0, 1]$ are floating-point values normalized by image width and height. Our custom tokenizer maps each coordinate to a single unified token, which preserves prompt conciseness in inputs and bypasses the multi-token autoregressive decoding traditionally required for coordinate serialization. Furthermore, treating coordinates as unified geometric units allows the tokenizer to flexibly support arbitrary pixel delimiter tokens (\emph{e.g.}, brackets, parentheses, or task-specific tags) without altering model weights.

\noindent\textbf{Coordinate Encoding via Visual Positional Embeddings.} 
To embed continuous coordinates $(x, y) \in [0, 1]^2$ into dense vectors that share the visual encoder's spatial frame, we reuse the pre-trained learned 2D positional embedding grid $\mathcal{P}_{\text{vis}} \in \mathbb{R}^{G \times G \times d_{\text{vis}}}$ of the Vision Transformer (e.g., $G = 48$). Rather than quantizing coordinates into discrete grid bins, we query $\mathcal{P}_{\text{vis}}$ via bilinear interpolation and project the resulting feature into the language hidden dimension $\mathbb{R}^{d_{\text{text}}}$ with a single linear projection (\cref{fig:semimoe_arch}f):
\begin{equation}
    \mathbf{e}_{\text{pix}}(x, y) = \mathbf{W}_{\text{em}} \, \mathcal{B}\big(\mathcal{P}_{\text{vis}}, (x, y)\big) + \mathbf{b}_{\text{em}},
    \label{eq:bilinear_pos}
\end{equation}
where $\mathcal{B}(\mathbf{F}, \mathbf{p})$ denotes the continuous 2D bilinear interpolation operator that samples a discrete grid $\mathbf{F}$ at normalized coordinate $\mathbf{p} \in [0, 1]^2$. As demonstrated in our ablation studies (\cref{sec:exp_ablations}), reusing the visual encoder's learned positional embeddings provides an essential geometric inductive bias that natively aligns coordinates with visual token representations and eases training.

\subsection{Differentiable ESDF Supervision for Pointing}
\label{sec:method_masksdf}

In real-world robotic manipulation, actionable affordances and placement regions span continuous 2D zones $M \subset [0, 1]^2$, forming an inherently \emph{multi-peak distribution} with multiple valid contact locations. Conventional point grounding enforces single-point regression (e.g., $\ell_1$ or Smooth L1 distance to an arbitrary mask centroid). This penalizes the model for predicting any valid contact mode away from the center and often forces it to output an arithmetic centroid that lands outside valid physical space. To resolve this without costly mask decoders, we propose ESDF supervision (\cref{fig:semimoe_arch}e).

\noindent\textbf{ESDF Formulation.} 
As illustrated in \cref{fig:semimoe_arch}e, for a target affordance region $M \subset [0, 1]^2$ enclosed by boundary contour $\partial M$, the signed distance $\phi_M(p)$ of any 2D point $p \in [0, 1]^2$ measures its Euclidean distance to the nearest boundary:
\begin{equation}
    \phi_M(p) = \begin{cases}
        - \min_{q \in \partial M} \|p - q\|_2, & \text{if } p \in M , \\
        + \min_{q \in \partial M} \|p - q\|_2, & \text{if } p \notin M ,
    \end{cases}
    \label{eq:sdf_definition}
\end{equation}
where $q \in \partial M$ represents a point on the mask boundary $\partial M$. Thus, $\phi_M(p) < 0$ inside the mask, $\phi_M(p) = 0$ on the boundary, and $\phi_M(p) > 0$ outside.

\noindent\textbf{Differentiable Bilinear Grid Sampling.} 
Given the predicted continuous coordinate $\hat{\mathbf{p}} = (\hat{x}, \hat{y}) \in [0, 1]^2$ emitted by the continuous coordinate head, we sample the scalar signed distance $\phi(\hat{\mathbf{p}})$ directly from the target ESDF map $\Phi \in \mathbb{R}^{S \times S}$ using the bilinear sampling operator $\mathcal{B}$:
\begin{equation}
    \phi(\hat{\mathbf{p}}) = \mathcal{B}(\Phi, \hat{\mathbf{p}}).
    \label{eq:grid_sample}
\end{equation}
Because the bilinear operator $\mathcal{B}$ is differentiable with respect to sampling coordinates, gradients $\frac{\partial \phi}{\partial \hat{\mathbf{p}}}$ flow directly into $\hat{\mathbf{p}}$, enabling end-to-end backpropagation without discrete approximations.

\noindent\textbf{Asymmetric Continuous Loss Objective.} 
We formulate an asymmetric loss $\mathcal{L}_{\text{sdf}}$ over the signed distance $\phi$:
\begin{equation}
    \mathcal{L}_{\text{sdf}}(\phi) = \begin{cases}
        0.2 \cdot \exp(25.0 \cdot \phi), & \text{if } \phi \le 0 , \\
        0.2 + 5.0 \cdot \phi + 8.0 \cdot \phi^2, & \text{if } \phi > 0 .
    \end{cases}
    \label{eq:sdf_loss}
\end{equation}
This formulation ensures $C^1$ smoothness at the boundary $\phi = 0$. For exterior points ($\phi > 0$), the linear-quadratic penalty exerts an increasingly strong inward pull as distance grows. Crucially, assigning a small, exponentially decaying loss to interior points ($\phi \le 0$) gently pushes predictions near the boundary further inward for a safer margin and more robust localization, while rapidly vanishing with depth to accommodate arbitrary interior contact peaks without center-seeking bias.

\noindent\textbf{Joint Training Objective.} 
The total multi-task objective combines language modeling, routing supervision, and coordinate loss:
\begin{equation}
    \mathcal{L} = \mathcal{L}_{\text{LM}} + \mathcal{L}_{\text{router}} + \lambda_{\text{coord}} \mathcal{L}_{\text{coord}} + \lambda_{\text{esdf}} \mathcal{L}_{\text{sdf}},
    \label{eq:total_loss}
\end{equation}
where $\mathcal{L}_{\text{coord}}$ (\cref{eq:coord_loss}) supervises fixed coordinate targets (trajectories and bounding boxes), $\mathcal{L}_{\text{sdf}}$ (\cref{eq:sdf_loss}) supervises regional affordance masks, and $\lambda_{\text{coord}}, \lambda_{\text{esdf}}$ balance the terms.

\subsection{Two-Stage Training Curriculum}
\label{sec:method_stages}

\noindent\textbf{Model Initialization.}
We instantiate \method at two scales: our 4B model is initialized from Qwen3.5-4B, and our 9B model is initialized from RynnBrain 1.1-9B~\cite{rynnbrain2024} (which itself is initialized from Qwen3.5-9B). Across the top $K=8$ transformer layers, the spatial expert $\mathrm{FFN}_{\text{pix}}$ is initialized by copying weights from the corresponding pre-trained $\mathrm{FFN}_{\text{lang}}$, whereas the router $\mathrm{MLP}_r$, pixel head $\mathrm{MLP}_p$, and coordinate projection $\mathbf{W}_{\text{em}}$ are randomly initialized. We refer to these four components collectively as the \emph{spatial modules}.

\noindent\textbf{Stage 1: Expert Cold-Start and Routing Warmup.} 
To prevent the randomly initialized modules from destabilizing the pre-trained backbone, we first freeze the vision encoder, merger, and token embeddings. We apply LoRA to the bottom Transformer layers (layers 0--3) while fully training the \emph{spatial modules} to stabilize routing boundaries and align continuous coordinate embeddings.

\noindent\textbf{Stage 2: Full Multi-Layer LoRA and Spatial Specialization.} 
In Stage 2, we expand LoRA adaptation across all language layers while continuing full fine-tuning of the \emph{spatial modules} over our full multimodal mixture, yielding our final \method model.

\section{Experiments}
\label{sec:experiments}

\begin{table}[t]
\centering
\footnotesize
\setlength{\tabcolsep}{4.5pt}
\renewcommand{\arraystretch}{1.18}
\begin{tabularx}{\columnwidth}{@{} l >{\raggedright\arraybackslash}X @{}}
\toprule
\textbf{Task Modality} & \textbf{Training Data Sources} \\
\midrule
\multicolumn{2}{@{}l}{\textit{\textbf{General Multimodal Cognition}}} \\
General Understanding & Visual7W~\cite{zhu2016visual7w}, Visual Genome~\cite{krishna2017visual} \\
Spatial Understanding & RefSpatial~\cite{zhou2025roborefer}, RoboSpatial~\cite{song2025robospatial}, RoboPoint~\cite{yuan2024robopoint} \\
General Localization & EgoObjects~\cite{zhu2023egoobjects}, LVIS~\cite{gupta2019lvis}, PixMo-Points~\cite{deitke2024molmo}, V3Det~\cite{wang2023v3det} \\
\midrule
\multicolumn{2}{@{}l}{\textit{\textbf{Embodied Physical Grounding \& Planning}}} \\
Embodied Understanding & FSD~\cite{yuan2026fromseeing}, Robo2VLM~\cite{chen2025robo2vlm} \\
Grounding \& Affordance & RAGNet~\cite{wu2025ragnet}, HOI4D~\cite{liu2022hoi4d}, RoboAfford~\cite{hao2025roboafford}, ShareRobot~\cite{ji2025robobrain}, GraspNet~\cite{fang2020graspnet}, VL-Grasp~\cite{lu2023vlgrasp}, GraspClutter6D~\cite{back2025graspclutter6d} \\
Trajectory Prediction & MolmoAct~\cite{lee2025molmoact}, $A_0$~\cite{xu2025a0} \\
Reasoning \& Planning & RoboVQA~\cite{sermanet2024robovqa}, ShareRobot~\cite{ji2025robobrain} \\
\midrule
\multicolumn{2}{@{}l}{\textbf{Total Multi-Task Mixture:} $\sim$\textbf{12.0 Million Training Samples}} \\
\bottomrule
\end{tabularx}
\caption{\textbf{Comprehensive multi-task training mixture for \method}. Datasets are structured across general multimodal cognition and embodied physical grounding. For composite suites (\emph{e.g.}, ShareRobot~\cite{ji2025robobrain}), we incorporate their respective domain subsets. }
\label{tab:data_mixture}
\end{table}

\begin{table*}[t]
\centering
\small
\setlength{\tabcolsep}{1.0pt}
\renewcommand{\arraystretch}{1.0}
\begin{tabularx}{\textwidth}{l|*{5}{Y}|*{6}{Y}}
\toprule
\textbf{Benchmark} & \footnotesize\textbf{Qwen3.5} & \footnotesize\textbf{Rynn1.0} & \footnotesize\textbf{RoboBrain} & \footnotesize\textbf{Ours} & \footnotesize\textbf{No-MoE} & \footnotesize\textbf{RoboBrain} & \footnotesize\textbf{Rynn1.1} & \footnotesize\textbf{Cosmos3} & \footnotesize\textbf{Embodied} & \footnotesize\textbf{HY} & \footnotesize\textbf{Ours} \\
& \footnotesize(4B) & \footnotesize(4B) & \footnotesize(4B) & \footnotesize(4B) & \footnotesize(4B) & \footnotesize(8B) & \footnotesize(9B) & \footnotesize(8B) & \footnotesize(8B) & \footnotesize(30A3B) & \footnotesize(9B) \\
\midrule
Part-Affordance & 17.9 & 15.5 & \second{34.8} & \first{57.2} & 47.0 & 25.1 & 42.9 & 32.2 & \first{82.6} & 63.7 & \second{82.2} \\
PIOBench & 49.4 & \first{58.5} & \second{56.8} & 52.1 & 50.2 & 60.2 & \second{62.2} & 54.4 & 61.6 & \first{65.3} & 60.7 \\
RefSpatial-Bench & 39.7 & \first{56.8} & 53.3 & \second{53.8} & 50.9 & 54.9 & \first{62.0} & 44.4 & 52.4 & 53.4 & \second{55.2} \\
RoboAfford & 65.6 & 73.1 & \first{77.4} & \second{73.5} & 71.0 & 72.5 & 72.8 & \first{76.2} & 60.6 & 71.5 & \second{74.8} \\
RoboRefit & 81.0 & 61.0 & \first{83.5} & \second{82.0} & 78.7 & \second{82.4} & 81.9 & 55.4 & 77.2 & \first{88.2} & 82.0 \\
VABench-Point & 37.8 & 2.3 & \second{45.8} & \first{46.7} & 50.3 & 9.6 & 13.2 & 45.2 & \second{61.4} & 59.7 & \first{62.7} \\
Where2Place & 58.7 & \second{70.9} & 39.3 & \first{71.9} & 66.0 & 69.0 & \second{72.7} & 71.0 & \first{73.0} & 65.0 & 72.6 \\
ShareRobot-Traj & 77.5 & 81.7 & \first{86.8} & \second{85.9} & 86.0 & \second{85.2} & 80.9 & 65.5 & 69.2 & 76.7 & \first{87.1} \\
VABench-Trace & 79.5 & \second{84.3} & 83.5 & \first{87.2} & 87.2 & 86.7 & 87.9 & 81.6 & \second{89.8} & 79.7 & \first{90.2} \\
\midrule
\textbf{Average} & 56.3 & 56.0 & \second{62.4} & \first{67.8} & 65.3 & 60.6 & 64.1 & 58.4 & \second{69.8} & 69.2 & \first{74.2} \\
\bottomrule
\end{tabularx}
\caption{\textbf{Comprehensive Evaluation on Coordinate-Based Spatial Grounding Tasks}. Scores are on a $0$--$100$ scale ($\uparrow$). \firstbox~and~\secondbox~denote the best and second-best results among parameter-comparable models, respectively. Model abbreviations: Rynn1.0 (RynnBrain1.0), Rynn1.1 (RynnBrain1.1), RoboBrain (RoboBrain2.5), Cosmos3 (Cosmos3-Nano), Embodied (Embodied-R1.5), HY (HY-Embodied-VLM-1.0). Note that No-MoE (Ours-4B without MoE) is an ablation baseline and is not marked by color highlights.}
\label{tab:spatial_benchmarks}
\vspace{2pt}
\begin{tabularx}{\textwidth}{l|*{5}{Y}|*{6}{Y}}
\toprule
\textbf{Benchmark} & \footnotesize\textbf{Qwen3.5} & \footnotesize\textbf{Rynn1.0} & \footnotesize\textbf{RoboBrain} & \footnotesize\textbf{Ours} & \footnotesize\textbf{No-MoE} & \footnotesize\textbf{RoboBrain} & \footnotesize\textbf{Rynn1.1} & \footnotesize\textbf{Cosmos3} & \footnotesize\textbf{Embodied} & \footnotesize\textbf{HY} & \footnotesize\textbf{Ours} \\
& \footnotesize(4B) & \footnotesize(4B) & \footnotesize(4B) & \footnotesize(4B) & \footnotesize(4B) & \footnotesize(8B) & \footnotesize(9B) & \footnotesize(8B) & \footnotesize(8B) & \footnotesize(30A3B) & \footnotesize(9B) \\
\midrule
BLINK & \second{78.8} & \first{82.2} & 78.0 & 75.5 & 76.0 & 80.9 & \second{85.0} & 82.4 & 77.8 & \first{87.3} & 80.6 \\
CV-Bench & 86.2 & \first{89.2} & \second{87.4} & 78.2 & 72.9 & 87.6 & 87.7 & 88.0 & 86.8 & \first{89.7} & \second{88.7} \\
EmbSpatial & \second{77.3} & \first{80.6} & 75.9 & 68.9 & 73.9 & 77.9 & \second{81.9} & 80.0 & 76.0 & \first{82.7} & 81.9 \\
MindCube & 35.0 & \first{55.5} & 34.0 & \second{46.2} & 45.1 & 29.3 & \first{81.1} & 32.8 & 27.9 & 70.0 & \second{79.0} \\
SAT & \second{69.3} & 65.3 & 64.0 & \first{71.3} & 67.3 & 66.0 & \second{76.7} & 54.0 & 60.7 & \first{78.0} & \first{78.0} \\
VSI-Bench & 53.0 & \first{65.7} & 41.5 & \second{56.1} & 43.8 & 44.1 & \first{66.0} & 50.4 & 59.2 & 58.9 & \second{65.3} \\
ViewSpatial & 39.7 & \first{54.5} & 41.6 & \second{46.2} & 45.8 & 39.1 & \first{55.2} & 52.0 & 43.7 & \second{53.3} & 52.1 \\
COSMOS & \first{60.8} & \second{58.2} & 53.9 & \first{60.8} & 58.0 & 58.4 & 64.5 & \second{67.1} & \first{68.6} & 66.9 & 62.9 \\
ERQA & \first{46.0} & \second{43.8} & 40.2 & 42.8 & 42.3 & 44.8 & 43.0 & 45.0 & 37.3 & \first{60.8} & \second{47.0} \\
ERQA-PLUS & \first{82.8} & 77.3 & \second{81.8} & 80.6 & 78.5 & 81.0 & \first{83.6} & \first{83.6} & 80.7 & \second{82.2} & 80.8 \\
RoboVQA & 45.4 & \second{61.9} & 38.9 & \first{62.9} & 63.6 & 43.7 & \second{60.9} & 50.2 & 60.8 & 41.2 & \first{63.6} \\
\midrule
\textbf{Average} & 61.3 & \first{66.7} & 57.9 & \second{62.7} & 60.7 & 59.3 & \first{71.4} & 62.3 & 61.8 & 70.1 & \second{70.9} \\
\bottomrule
\end{tabularx}
\caption{\textbf{Evaluation on Text-Based Embodied Understanding \& Reasoning Tasks}. \firstbox~and~\secondbox~denote the best and second-best results among parameter-comparable models, respectively.}
\label{tab:reasoning_benchmarks}
\end{table*}


\subsection{Experimental Setup}
\label{sec:exp_setup}

\noindent\textbf{Training Data Preparation.}
As summarized in \cref{tab:data_mixture}, we curate a balanced 12M-sample training mixture spanning general multimodal cognition and embodied physical grounding. For object-pointing datasets with only point or box labels, we prompt SAM3~\cite{carion2026sam} to extract masks of irregular objects and convert them into ESDF maps for Mask-SDF supervision (\cref{sec:method_masksdf}), while part-level affordances use dedicated datasets.

\noindent\textbf{Evaluation Benchmarks.}
Our evaluation spans two complementary domains:
\begin{enumerate}[leftmargin=*,noitemsep,topsep=1pt]
    \item \emph{Coordinate-Based Spatial Grounding Tasks (9 Benchmarks)}: Tasks that require the model to output 2D coordinates in image plane. This suite covers affordance and interaction pointing on Part-Affordance-2K~\cite{yuan2025embodiedr1}, RoboAfford~\cite{hao2025roboafford}, VABench-Point~\cite{yuan2026fromseeing}, and Where2Place~\cite{yuan2024robopoint}; referring and spatial localization on PIOBench~\cite{xue2025pointitout}, RefSpatial-Bench~\cite{zhou2025roborefer}, and RoboRefit~\cite{lu2023vlgrasp}; and trajectory prediction on ShareRobot-Trajectory~\cite{ji2025robobrain} and VABench-Visual-Trace~\cite{yuan2026fromseeing}.
    
    \item \emph{Text-Based Embodied Understanding \& Reasoning Tasks (11 Benchmarks)}: Tasks where the model responds in natural language, such as multiple-choice answers, numerical estimates, or textual action plans. These benchmarks encompass scene spatial perception and relational QA on BLINK~\cite{fu2024blink}, CV-Bench~\cite{tong2024cambrian}, EmbSpatial~\cite{du2024embspatial}, and SAT~\cite{ray2025sat}; multi-view and video spatio-temporal reasoning on MindCube~\cite{wang2026mindcube}, ViewSpatial~\cite{li2025viewspatial}, and VSI-Bench~\cite{yang2025thinking}; and physical dynamics and embodied task planning on COSMOS~\cite{azzolini2025cosmos}, ERQA~\cite{geminirobotics2025}, ERQA-PLUS~\cite{yang2026erqaplus}, and RoboVQA~\cite{sermanet2024robovqa}.
\end{enumerate}

\begin{figure*}[t]
    \centering
    \includegraphics[width=\textwidth]{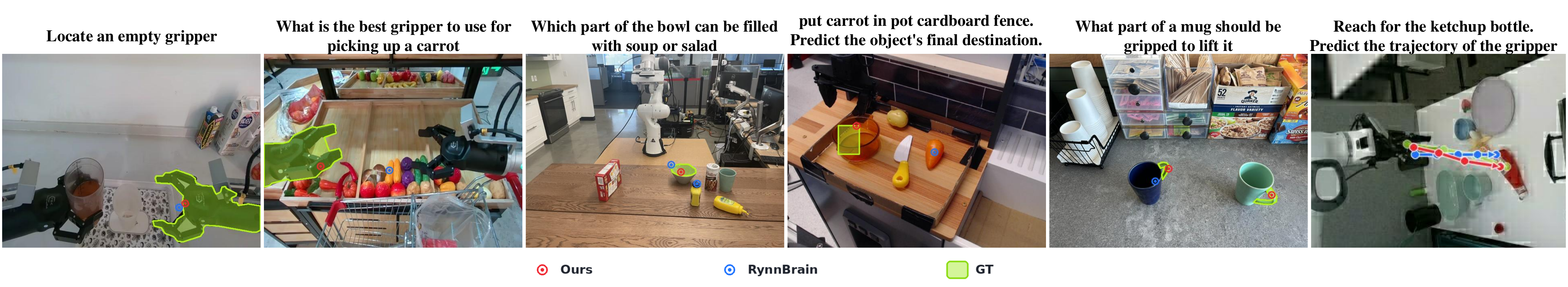}
    \caption{\textbf{Qualitative Comparison against RynnBrain1.1-9B.} \method (red) accurately hits ground-truth regions (green) and tracks target trajectories, while the base model (blue) drifts or misses fine-grained parts.}
    \label{fig:qualitative}
\end{figure*}
Because \method directly outputs complete, continuous 2D coordinates rather than character-by-character token strings, it does not require verbose prompt constraints specifying coordinate output formatting. Accordingly, we omit verbose format instructions on coordinate-based tasks and simply append \texttt{Type: point/bbox/trajectory} to specify the target output for our model.

\noindent\textbf{Metrics.}
All scores are reported on a standard $0$--$100$ scale (higher is better). 
Point grounding tasks are evaluated using micro-averaged $\mathrm{F}_1 = 2\mathrm{P}_{\mu}\mathrm{R}_{\mu}/(\mathrm{P}_{\mu}+\mathrm{R}_{\mu})\times 100$, where precision $\mathrm{P}_{\mu} = \sum_i H_i / \sum_i P_i$ measures the proportion of predicted points falling inside valid ground-truth masks, and recall $\mathrm{R}_{\mu} = \sum_i \min(M_i, G_i) / \sum_i G_i$ measures the proportion of covered targets ($P_i, G_i$ are prediction and target counts; $H_i, M_i$ are hit and matched counts for sample $i$)
For trajectory benchmarks (ShareRobot-Trajectory and VABench-Visual-Trace), following standard evaluation protocols, trajectory accuracy is computed as $100 \times (1 - \text{mean\_RMSE} / 1000)$. The best and second-best scores among parameter-comparable models are highlighted in dark blue (\firstbox) and light blue (\secondbox), respectively.

\noindent\textbf{Baselines.}
We evaluate \method at both 4B and 9B scales against representative open-source and published embodied foundation models. At the 4B scale, we compare with the base model Qwen3.5 (4B), RynnBrain 1.0 (4B)~\cite{rynnbrain2024}, and RoboBrain 2.5 (4B)~\cite{robobrain2024}. At the 8B--9B (and MoE) scale, our competitors include RoboBrain 2.5 (8B)~\cite{robobrain2024}, RynnBrain 1.1 (9B)~\cite{rynnbrain2024}, Cosmos3-Nano (8B)~\cite{agarwal2026cosmos}, Embodied-R1.5 (8B)~\cite{embodiedr15_2024}, and the large-scale mixture-of-experts model HY-Embodied-VLM-1.0 (30A3B)~\cite{hyembodied2024}. Note that some results for Embodied-R1.5, Cosmos3-Nano, and HY-Embodied-VLM in \cref{tab:spatial_benchmarks,tab:reasoning_benchmarks} are adopted from \emph{HY-Embodied-VLM-1.0}~\cite{hyembodied2024}, while all other models are evaluated locally under unified protocols.

\subsection{Main Results and Analysis}
\label{sec:exp_main_results}

\cref{tab:spatial_benchmarks} and \cref{tab:reasoning_benchmarks} evaluate \method across 9 coordinate-based spatial grounding tasks and 11 text-based embodied understanding and reasoning tasks, validating two key properties of our framework:

\noindent\textbf{1. Strong Gains on Coordinate-Based Spatial Grounding.}
As shown in \cref{tab:spatial_benchmarks}, \method substantially improves over its base models at both scales—rising from $56.3$ to 67.8 on 4B (over Qwen3.5-4B) and from $64.1$ to 74.2 on 9B ($+15.7 \%$ relative over RynnBrain 1.1)—while also outperforming specialized embodied VLMs including the $3\times$ larger 30B HY-Embodied-VLM. The advantage of our coordinate encoding is most evident on trajectory prediction, the most pixel-context-rich task: each waypoint is predicted conditioned on previously generated ones, so the model must read its own coordinates back as context. Since each coordinate is a single token embedded in the visual encoder's positional space, earlier waypoints stay aligned with image patches instead of being fragmented into digit tokens. Accordingly, \method-9B ranks first on both ShareRobot-Traj  and VABench-Trace.

\noindent\textbf{2. Preserving Text-Based Reasoning.}
Comparing \cref{tab:spatial_benchmarks} and \cref{tab:reasoning_benchmarks} relative to each model's initialization backbone demonstrates that our \method architecture prevents coordinate regression from interfering with pre-trained linguistic representations. By sharing lower layers for multimodal context while isolating spatial regression gradients within $\text{FFN}_{\text{pix}}$ across the top $K$ layers, \method achieves substantial gains in coordinate grounding ($+11.5$ on 4B and $+10.1$ on 9B) while maintaining text-based reasoning performance on par with its base models: \method-4B slightly improves over Qwen3.5-4B ($62.7$ \emph{vs.}$61.3$), and \method-9B closely matches RynnBrain 1.1 ($70.9$ \emph{vs.} $71.4$).

\noindent\textbf{3. Qualitative Comparison.}
\cref{fig:qualitative} shows qualitative results of \method-9B and RynnBrain1.1-9B across evaluated benchmarks: \method hits valid object/part affordance regions (including multi-peak mug handles) and accurate trajectory endpoints, whereas the base model often drifts onto background or off-target regions.

\subsection{Ablation Studies}
\label{sec:exp_ablations}

To validate our core design choices, we conduct ablation analyses on \method-4B, investigating continuous coordinate token encoding, mask-based supervision objectives, and modality-decoupled MoE routing.

\noindent\textbf{1. Coordinate Token Encoding Method.}
We initially considered three candidate schemes for encoding a coordinate token $(x, y) \in [0, 1]^2$: reusing the visual encoder's learned positional embeddings ($\mathbf{e}_{\text{pe}}$), sinusoidal Fourier frequency encoding ($\mathbf{e}_{\text{f}}$), and local visual patch features sampled at that coordinate ($\mathbf{e}_{\text{i}}$). To allow the model to automatically learn which encoding scheme is most suitable, we combine the three candidates via a learnable gated fusion layer during training:
\begin{equation}
\begin{aligned}
    \tilde{\mathbf{e}}_{\text{fused}} &= w_{\text{pe}} \mathbf{e}_{\text{pe}} + w_{\text{f}} \mathbf{e}_{\text{f}} + w_{\text{i}} \mathbf{e}_{\text{i}}, \\
    \mathbf{e}_{\text{fused}} &= \frac{\|\mathbf{e}_{\text{pe}}\|_2}{\|\tilde{\mathbf{e}}_{\text{fused}}\|_2} \tilde{\mathbf{e}}_{\text{fused}},
\end{aligned}
\label{eq:pixel_fusion}
\end{equation}
where $\mathbf{e}_{\text{pe}}$ projects the interpolated visual positional embedding $\mathcal{P}_{\text{vis}}$ (\cref{eq:bilinear_pos}); $\mathbf{e}_{\text{f}}$ maps $12$-frequency sinusoidal Fourier features of $(x, y)$ (frequencies $2^k \pi$ for $k = 0, \dots, 11$) through a two-layer MLP; and $\mathbf{e}_{\text{i}}$ applies an affine-free LayerNorm and linear projection to pre-merger visual patch features sampled at $(x, y)$. 


Analyzing the learned gating weights and branch norm shares ($\|w_k \mathbf{e}_k\|_2 / \sum_{j} \|w_j \mathbf{e}_j\|_2$) at convergence reveals distinct behaviors:
\begin{itemize}[leftmargin=*,noitemsep,topsep=1pt]
    \item Learned Visual PE Dominates ($>97\%$): Reusing the visual encoder's learned PE accounts for over $97\%$ of the representation magnitude, as $\mathcal{P}_{\text{vis}}$ inherently encodes 2D metric geometry aligned with visual tokens.
    \item Image Features Provide Marginal Context ($\sim 3\%$): Pre-merger visual patch features contribute only $\sim 3\%$; while offering semantic appearance cues, patch-level features lack the fine spatial resolution required for exact coordinate localization.
    \item Fourier Encoding Fails to Contribute ($<1\%$): The learned Fourier weight vanishes to less than $1\%$ ($w_{\text{f}} \approx 0$), as generic sinusoidal encodings operate in an unaligned domain disjoint from the visual representation manifold.
\end{itemize}
These results confirm that directly reusing the visual encoder's learned PE provides the most effective coordinate representation, rendering auxiliary Fourier embeddings and patch feature injections unnecessary.


\begin{table}[t]
\centering
\small
\setlength{\tabcolsep}{6pt}
\renewcommand{\arraystretch}{1.12}
\resizebox{\columnwidth}{!}{
\begin{tabular}{l|cccc}
\toprule
\textbf{Supervision Objective} & \textbf{Part-Afford.} & \textbf{RoboRefit} & \textbf{ShareRobot-Traj} & \textbf{RoboAfford} \\
\midrule
(a) Without Mask (Centroid $\ell_1$) & 55.5 & 79.2 & 85.5 & 72.3 \\
(b) With Mask-SDF (Ours) & \textbf{57.2} & \textbf{82.0} & \textbf{85.9} & \textbf{73.5} \\
\bottomrule
\end{tabular}
}
\caption{\textbf{Ablation Study on Supervision Objectives (Evaluated on 4B Model)}. Comparison of single-point centroid regression versus our differentiable Mask-SDF supervision across affordance grounding, referring expressions, and trajectory prediction.}
\label{tab:ablations}
\end{table}

\noindent\textbf{2. ESDF Supervision on Pointing.}
Under identical training data and architecture, \cref{tab:ablations} compares single-point centroid regression ($\ell_1$) against our differentiable ESDF supervision. By accommodating arbitrary valid contact points on multi-peak affordances and irregular geometries rather than forcing predictions toward a single centroid, ESDF supervision consistently improves performance across affordance grounding and referring localization. Furthermore, ShareRobot-Traj also benefits due to more accurate localization of trajectory start and target points.

\noindent\textbf{3. Modality-Decoupled MoE.}
To isolate the effect of architectural decoupling, \cref{tab:spatial_benchmarks,tab:reasoning_benchmarks} compare \method-4B against a dense ablation (\textbf{No-MoE}) without top-layer FFN separation trained with the same training strategy, data mixture, and continuous coordinate head—retaining a final-layer router between the LM and pixel heads. Although No-MoE achieves competitive spatial grounding, updating shared FFN parameters under heterogeneous objectives induces severe cross-modal interference and mode drift. In particular, No-MoE exhibits a pronounced degradation on VSI-Bench, where unfamiliar question formats absent from the fine-tuning distribution frequently trigger spurious coordinate outputs. By confining spatial regression gradients to $\text{FFN}_{\text{pix}}$, our modality-decoupled MoE alleviates sensitivity to training data proportions, preserving the pre-trained VLM's general reasoning capabilities while further elevating spatial grounding accuracy.

\subsection{Inference Efficiency}
\label{sec:exp_latency}

\begin{table}[t]
\centering
\small
\setlength{\tabcolsep}{5pt}
\renewcommand{\arraystretch}{1.12}
\resizebox{\columnwidth}{!}{
\begin{tabular}{lcc}
\toprule
\textbf{Model} & \textbf{Coord. Decoding} & \textbf{Throughput (samples/s)} $\uparrow$ \\
\midrule
RynnBrain1.0-4B~\cite{rynnbrain2024} & Autoregressive Tokens & 5.28 \\
RoboBrain2.5-4B~\cite{robobrain2024} & Autoregressive Tokens & 3.41 \\
RynnBrain1.1-9B~\cite{rynnbrain2024} & Autoregressive Tokens & 2.61 \\
EmbodiedR1.5-8B~\cite{embodiedr15_2024} & Autoregressive Tokens & 2.81 \\
RoboBrain2.5-8B~\cite{robobrain2024} & Autoregressive Tokens & 1.58 \\
\midrule
\textbf{Ours-4B} & Direct Regression & \textbf{31.88 ($6.04\times$)} \\
\textbf{Ours-9B} & Direct Regression & \textbf{23.34 ($8.31\times$)} \\
\bottomrule
\end{tabular}
}
\caption{\textbf{Coordinate Inference Speed Comparison}. Measured in samples/second ($\text{samples/s} \uparrow$). Speedups in parentheses are computed conservatively relative to the fastest competing baseline at each model scale.}
\label{tab:latency}
\end{table}

To evaluate efficiency for spatial grounding in robotic tasks, we benchmark coordinate inference throughput on an NVIDIA RTX 5090 GPU. We construct an evaluation suite of 1,000 instances uniformly sampled across point affordance, referring localization, and motion trajectory benchmarks. Here, the evaluation metric is measured in samples per second (samples/s), where a complete coordinate tuple alongside its formatting delimiters constitutes one sample.

As shown in \cref{tab:latency}, conventional autoregressive VLMs serialize each 2D coordinate into 8 to 10 character and delimiter tokens, requiring iterative decoding passes that bottleneck throughput to only 1.58--5.28\,samples/s. In contrast, \method represents each coordinate as a single token and emits it in a single decoding pass via the coordinate head, completely bypassing multi-token autoregressive serialization. Consequently, \method-4B achieves 31.88 samples/s ($6.04\times$ speedup over RynnBrain1.0-4B), while \method-9B attains 23.34 samples/s ($8.31\times$ speedup over EmbodiedR1.5-8B), providing the responsiveness necessary for real-time robotic control.

\section{Conclusion}
\label{sec:conclusion}

In this paper, we introduced \method, an embodied vision-language model that harmonizes continuous physical manipulation with high-level multimodal reasoning. By decoupling top-layer FFNs while preserving shared self-attention, \method eliminates gradient interference between categorical text generation and geometric regression. Furthermore, by reusing pre-trained visual positional embeddings for continuous coordinate encoding and formulating differentiable Mask-SDF supervision for multi-peak affordances, our model achieves fast, accurate spatial grounding without auxiliary mask decoders. Extensive evaluations across 20 benchmarks validate that \method achieves state-of-the-art trajectory accuracy and outstanding affordance grounding while preserving conversational reasoning fidelity.

\section*{LLM Usage Statement}
We used Gemini solely for language polishing, grammar correction, and improving the readability of the manuscript, and OpenAI Codex to assist with auxiliary code implementation and debugging. All generated text and code were thoroughly reviewed, tested, and verified by the authors.

{
    \small
    \bibliographystyle{ieeenat_fullname}
    \bibliography{main}
}


\end{document}